\documentclass{article}

\PassOptionsToPackage{numbers, compress}{natbib}

\usepackage[preprint]{neurips_2024}

\usepackage[utf8]{inputenc} 
\usepackage[T1]{fontenc}    
\usepackage{hyperref}       
\usepackage{url}            
\usepackage{booktabs}       
\usepackage{amsfonts}       
\usepackage{nicefrac}       
\usepackage{microtype}      
\usepackage{xcolor}  
\usepackage[pdftex]{graphicx}

\title{Multi-Species Object Detection in Drone Imagery for Population Monitoring of Endangered Animals
}

\author{%
  Sowmya Sankaran \\
  Albuquerque Academy\\
  Albuquerque, New Mexico, 87109 \\
  \texttt{sans270@aa.edu} \\
}

\begin{document}

\maketitle

\begin{abstract}
  Animal populations worldwide are rapidly declining, and a technology that can accurately count endangered species could be vital for monitoring population changes over several years. This research focused on fine-tuning object detection models for drone images to create accurate counts of animal species. Hundreds of images taken using a drone and large, openly available drone-image datasets were used to fine-tune machine learning models with the baseline YOLOv8 architecture. We trained 30 different models, with the largest having 43.7 million parameters and 365 layers, and used hyperparameter tuning and data augmentation techniques to improve accuracy. While the state-of-the-art YOLOv8 baseline had only 0.7\% accuracy on a dataset of safari animals, our models had 95\% accuracy on the same dataset. Finally, we deployed the models on the Jetson Orin Nano for demonstration of low-power real-time species detection for easy inference on drones.

\end{abstract}

\section{Introduction}
In recent years, wildlife populations have been rapidly decreasing worldwide, with a 69\% decrease since 1970 \cite{decreasepercentage}. A study of over 71,000 species shows that 48\% of species have declining populations \cite{finn2023more}. Current counts of these species use approximate techniques such as core counts \cite{pollitt2003wetland}, point counts \cite{bibby1998bird}, and bird banding \cite{youngflesh2023demographic}, all of which involve human volunteers \cite{approximatecounts}, leading to difficulties in monitoring long-term population changes. Drone-based object detection could be used to accurately count animals. However, state-of-the-art object detection models are inaccurate on drone imagery.

Models such as YOLOv8 \cite{yolov8} and VGG16 \cite{simonyan2014very} are very accurate on several image-related tasks, but object detection on drone imagery has unique requirements, such as: (1) identifying fine features of small animals on images taken from afar; (2) handling high-resolution images, and (3) identifying species when they blend in to their environments. While there has been some previous work on adapting YOLOv8 for drone imagery, \cite{zhang2023drone} the improvements over the baseline YOLOv8 are a few percentage points. We fine-tune the YOLOv8 model for multi-species drone images, demonstrating up to 98.2\% accuracy. We deploy our model on the NVIDIA Jetson Orin Nano for a low-power hardware solution for real-time species detection. Additional information about this project can be found online \footnote{Slides: \url{https://tinyurl.com/2wvbs4x2} ; Talk: \url{https://youtu.be/XuLLuCdNegA}}

\textbf{Societal Impact:} This project was inspired by rapidly declining populations of migratory birds such as Sandhill cranes in our home state of New Mexico. The technology created in this project could play a vital role in determining changes in these populations. Our solution is both accurate and cost-effective, only costing \$800 including the Jetson Orin Nano and the drone.
Our approach can be used by park rangers worldwide to monitor endangered species populations in low-resource wildlife refuges even without a large budget for population monitoring.

To summarize, the main contributions of this research are:
\begin{itemize}
    \item Fine-tuning the YOLOv8 architecture on multi-species drone imagery for up to 135x increase in accuracy compared to the baseline YOLOv8.
    \item Comprehensive hyperparameter optimization with 30 different models to achieve up to 98.2\% accuracy when identifying datasets containing ten different species in drone imagery.
    \item Deploying the model on a low-power edge GPU for real-time object detection.

\end{itemize}

\section{Related Work}

YOLOv8 is version 8 in a family of continuously improving models based on the original YOLO architecture \cite{redmon2016you}. YOLO-NAS \cite{supergradients} improves on YOLOv8 using a Neural Architecture Search. It uses a quantization-friendly building block, which enables both high accuracy and lower latency. 
Both of these architectures have been used for many well-known object detection tasks. 

Norouzzadeh et al. \cite{norouzzadeh2018automatically} demonstrates identifying animal species in images from a camera-trap using convolutional neural networks. This study involves using five different deep learning architectures, including VGG and ResNet, to study the behavior of these animals. Our work is a small-scale, drone-image focused variation of this previous research. 


Recent work on drone-based object detection includes Drone-YOLO \cite{zhang2023drone} and YOLODrone \cite{sahin2021yolodrone}, which adapt the YOLOv8 and YOLOv3 architectures, respectively, for aerial imagery of urban environments. They primarily use the VisDrone2019 dataset. There is other recent work that uses drone imagery for identifying either large mammals \cite{lenzi2023artificial} or specific bird species \cite{francis2020counting}. \citet{corcoran2021automated} surveys several of these approaches. While YOLOv8 has been used for several object detection tasks, to our knowledge there has been no work that has focused on the challenging task of detecting \textit{multiple small animal species} in aerial imagery, mixing mammals and birds. %






\section{Methodology}
\begin{figure}
    \centering
    \includegraphics[width=1\linewidth]{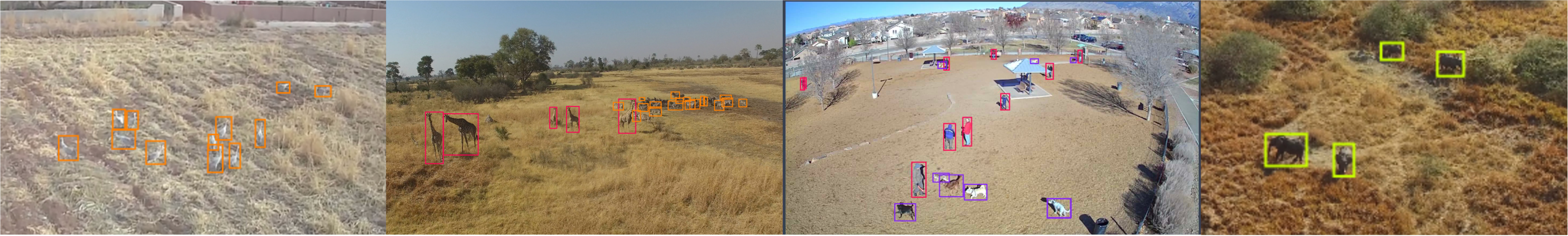}
    \caption{From left, images of: Sandhill cranes (taken by us); giraffes and zebras \cite{elephantgiraffe}; dogs and humans (taken by us); and elephants {\cite{elephant}.}}
    \label{fig:labeledimages}
\end{figure}

\begin{table}
  \caption{The datasets used for model training, with the number of images in the training, validation, and testing sets.}
  \label{datasettable}
  \centering
  \begin{tabular}{llll}
    \toprule
    Dataset     & Train     & Valid & Test \\
    \midrule
    Baseline (Sandhill crane, goose, dog, human) & 147  & 47 & 22     \\
    Image-Augmented (Sandhill crane, goose, dog, human)     & 267 & 47 & 22      \\
    Bounding Box-Augmented (Sandhill crane, goose, dog, human)     & 440  & 47 & 22  \\
     Elephant Dataset (elephant)     & 1015  & 328 & 152   \\
    Safari Dataset (elephant, giraffe, impala, lechwe, tsessebe, zebra)  & 2782  & 905 & 427\\
    \bottomrule
  \end{tabular}
\end{table}

Our methodology followed the entire machine learning workflow, from data collection, preparation, and augmentation to training, fine-tuning, and evaluating the models. In order to collect data of migratory birds, we captured hundreds of images using a drone at sites around Albuquerque, New Mexico. The images we collected included Sandhill cranes and Canada geese. We use dogs as a substitute for other mammals such as coyotes or wolves, and include humans as they are present in many images with dogs (Figure \ref{fig:labeledimages}). A representative selection of these images -- captured from different angles, environments, and times of day -- was used to create the baseline dataset (Table \ref{datasettable}). Each of these images was hand-labeled with bounding boxes indicating their species using Roboflow. Additionally, thousands of openly available drone images from African wildlife refuges (Figure \ref{fig:labeledimages}) were used to assemble two datasets, containing six different species (Table \ref{datasettable}). The distribution of the images in the datasets for training, validation, and testing was 68\%, 22\%, and 10\%, respectively.

\begin{figure}
    \centering
    \includegraphics[width=1\linewidth]{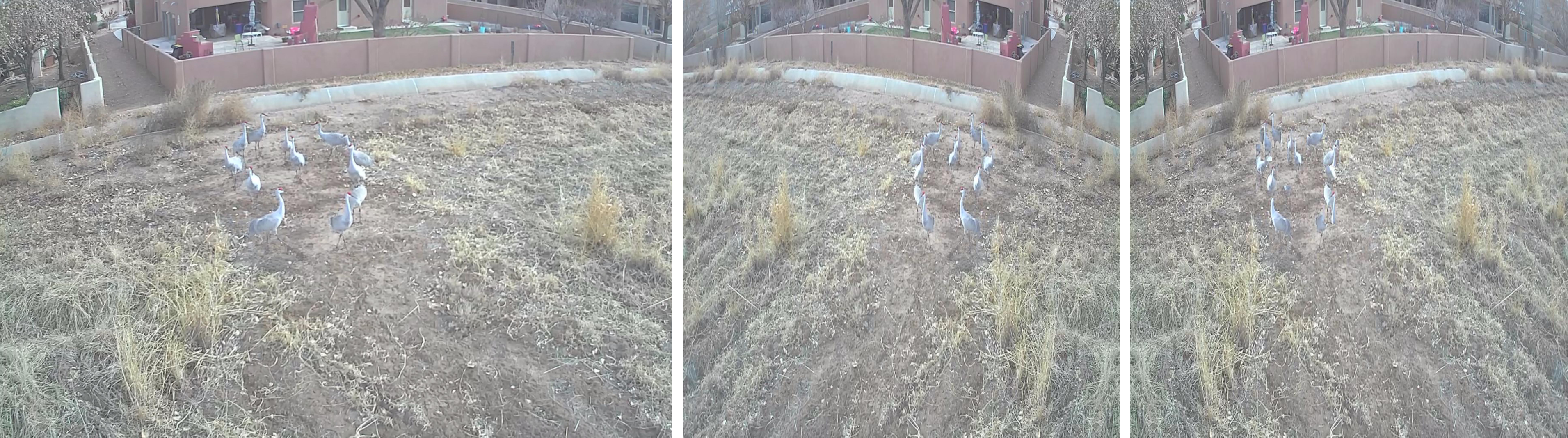}
    \caption{Examples of the data augmentation techniques, from left: original image of Sandhill cranes; image with resize and vertical-axis flip; image with resize and bounding box flip.}
    \label{fig:aug}
\end{figure}

Several data augmentation techniques (Figure \ref{fig:aug}) were used to create additional training examples for the models. These included a vertical axis image flip and a vertical axis bounding box flip. We also resized each image to $1024\times1024$ in order to optimize training efficiency. These data augmentation techniques had marginal impact on accuracy. We fine-tuned the Nano, Small, Medium, and Large YOLOv8 model sizes, with number of parameters ranging from 3.2 million to 43.7 million. We used Pytorch in Google Colab on NVIDIA A100 and V100 GPUs to train these models. 

In order to improve accuracy, we did a comprehensive hyperparameter search. The adjusted hyperparameters included the \textit{model size}, \textit{image size} ($640\times640$ to $1024\times1024$), \textit{number of epochs} (60-700), \textit{learning rate} (0.01-0.001), and \textit{patience} (25-300). The final step in the process was deploying the model on hardware for real-time species detection. We used the NVIDIA Jetson Orin Nano, an embedded computing edge device, for inference.

\section{Experimental Results}

The trained models were evaluated using several metrics, such as the mean average precision calculated over the intersection over union (IOU) threshold of 0.50 (\textbf{mAP-50}) and 0.50 to 0.95 (\textbf{mAP-50-95}), for all classes. The highest mAP-50 we achieved was 98.2 (model trained on all ten species and tested on elephants), and the highest mAP-50-95 we achieved was 66.3 (model trained and tested on elephants). We also evaluated the models using \textbf{precision} and \textbf{recall}. The highest values were achieved on the elephant dataset, with 97.1 precision and 97.4 recall.

\begin{figure}
    \centering
    \includegraphics[width=1.0\linewidth]{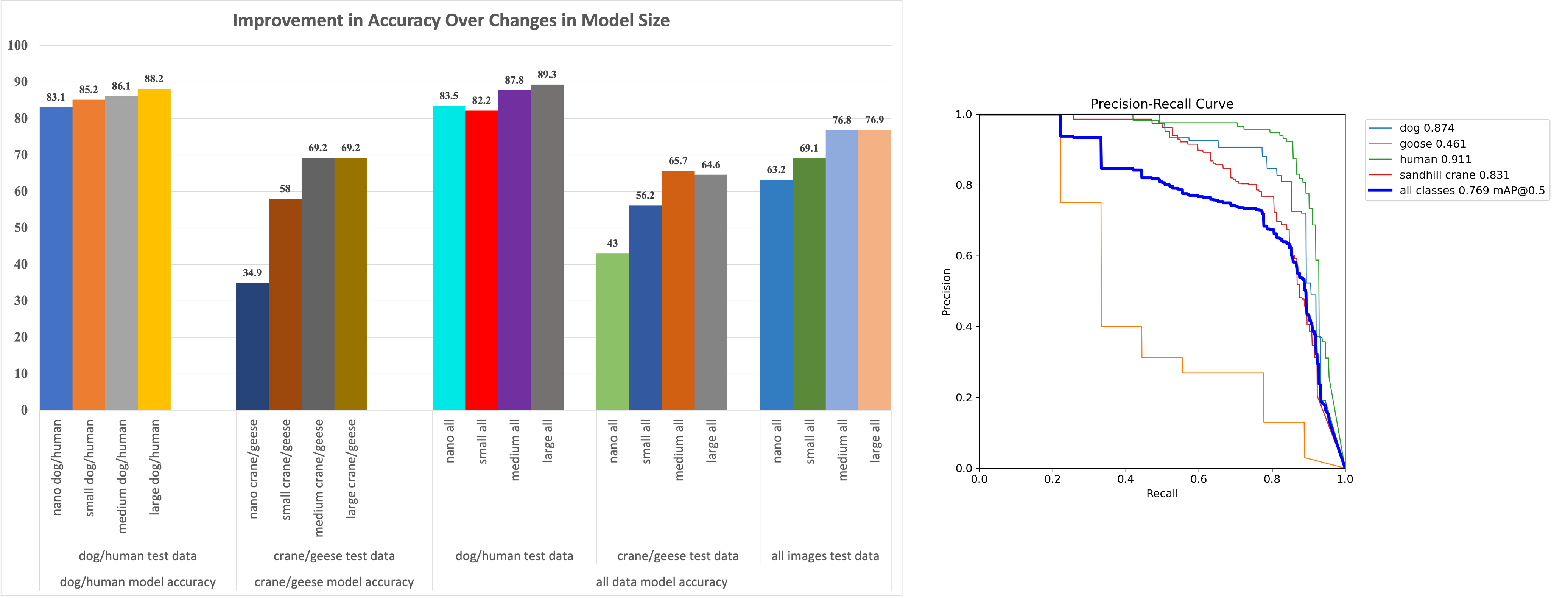}
    \caption{Left, the model accuracy for 20 different models, with 4 model sizes trained on the baseline datasets. The 20 models are grouped in a hierarchical fashion, showing training and testing data. Right, a precision-recall curve for a model with 89.3\% accuracy. }
    \label{fig:hypgraph}
\end{figure}

In total, we trained 30 different models, adjusting the hyperparameters to improve accuracy. We ensured that the datasets included a combination of several species to avoid overfitting to one species. The testing included 59 separate inference runs on different datasets. Figure \ref{fig:hypgraph} shows the accuracy for 20 different models, four different model sizes, and a combination of three sets of training data and three sets of testing data. In general, the medium and large model sizes are the most accurate. The precision-recall curve for a model that includes four species is also shown. The accuracy for the geese tends to be lower because of limited training data. All datasets, training logs, and inference results are publicly available \footnote{\url{https://tinyurl.com/bdhwen8d}}.

\begin{figure}
    \centering
    \includegraphics[width=1\linewidth]{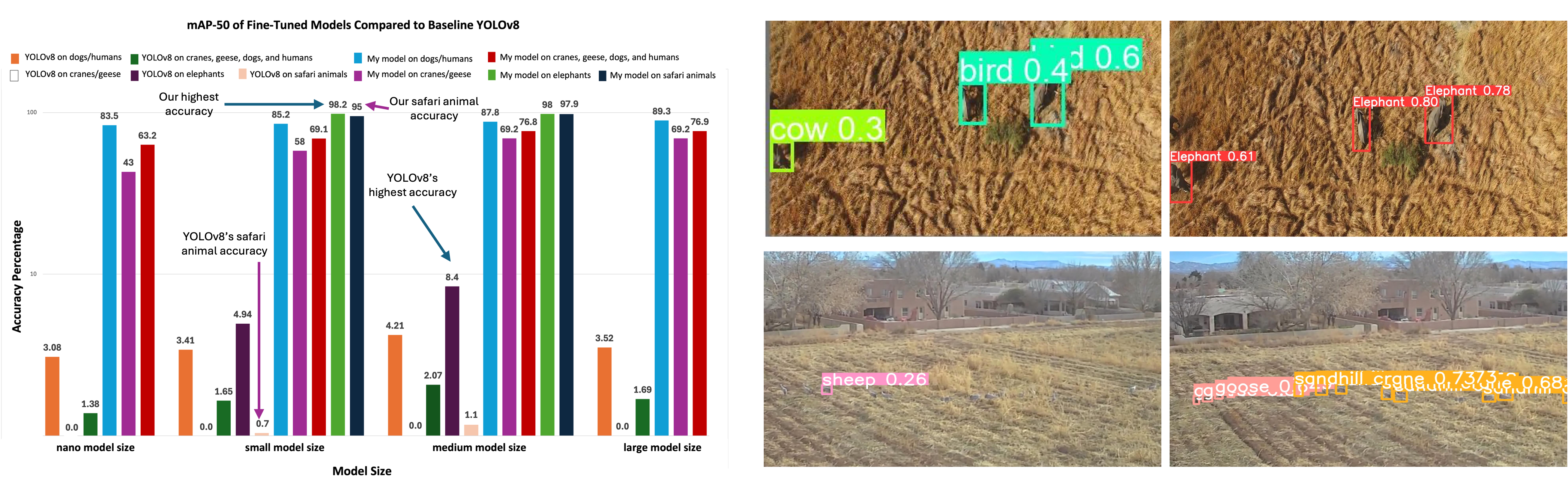}
    \caption{Left, a logarithmic-scale graph of YOLOv8's accuracy on datasets compared to our model's accuracy. Right, YOLOv8’s predictions on two images are on the left and our model’s predictions are on the right. YOLOv8 could not identify the elephants or the bird species.}
    \label{fig:yolocompare}
\end{figure}

As a next step, we compared our best models against the baseline YOLOv8. For these evaluations, we used the models trained on all datasets, and ran inference on individual species. Our highest accuracy was 98.2\%, achieved when the model was tested on elephant data (Figure \ref{fig:yolocompare}, blue arrows). On the same dataset, YOLOv8's accuracy was 8.4\%, which was the baseline's highest accuracy on any dataset. Cranes and geese were the most challenging species to identify. While our models were able to successfully identify 74.7\% of cranes and geese, YOLOv8 was never able to identify a single bird, resulting in 0.0\% accuracy. Our models had very high accuracy on the safari animals, as this was a very large dataset with thousands of images, achieving 95\% accuracy. As YOLOv8 could only achieve 0.7\% accuracy on the same dataset, our models achieved 135x the baseline's accuracy (Figure \ref{fig:yolocompare}, purple arrows). We also attempted to use YOLO-NAS \cite{supergradients} as our baseline. However, we observed better accuracy with YOLOv8.

Finally, we deployed the models on the Jetson Orin Nano and evaluated inference speed. We achieved a trimmed mean of 24.38 milliseconds inference time per image, which translates to our model being able to identify species in 40 fps videos. We clocked the cores at their maximum frequency to achieve a 4.42\% improvement in inference time.

\section{Conclusion and Societal Impact}

We demonstrated a way to accurately count animal species in drone imagery through fine-tuning YOLOv8. Our models outperformed state-of-the-art models such as the YOLOv8 baseline, achieving up to 135x better accuracy. The results show the importance of hyperparameter optimization and sufficiently large datasets to achieve high accuracy. We also demonstrated that the YOLOv8 architecture can be used for the complex task of identifying multiple species in aerial imagery.


In the future, we would like to expand our datasets to contain even more species based on ever-evolving drone datasets. Additionally, having a better GPU capability would help train larger models. We would also like to develop a fully automated detection system and deploy our technology on a drone. This research provides a number of opportunities for studying animal behavior \cite{schad2023opportunities} and could benefit wildlife refuges and park rangers worldwide. Our models offer a path for accurate, automated counting and monitoring of species with deteriorating populations using images taken from drones.


\bibliographystyle{plainnat}
\bibliography{main}

\begin{thebibliography}{19}
\providecommand{\natexlab}[1]{#1}
\providecommand{\url}[1]{\texttt{#1}}
\expandafter\ifx\csname urlstyle\endcsname\relax
  \providecommand{\doi}[1]{doi: #1}\else
  \providecommand{\doi}{doi: \begingroup \urlstyle{rm}\Url}\fi

\bibitem[Aharon et~al.(2021)Aharon, {Louis-Dupont}, {Ofri Masad}, Yurkova,
  {Lotem Fridman}, {Lkdci}, Khvedchenya, Rubin, Bagrov, Tymchenko, Keren,
  Zhilko, and {Eran-Deci}]{supergradients}
Shay Aharon, {Louis-Dupont}, {Ofri Masad}, Kate Yurkova, {Lotem Fridman},
  {Lkdci}, Eugene Khvedchenya, Ran Rubin, Natan Bagrov, Borys Tymchenko, Tomer
  Keren, Alexander Zhilko, and {Eran-Deci}.
\newblock Super-gradients, 2021.
\newblock URL \url{https://zenodo.org/record/7789328}.

\bibitem[Almond et~al.(2022)Almond, Grooten, Juffe~Bignoli, and
  Petersen]{decreasepercentage}
R.E.A. Almond, M.~Grooten, D.~Juffe~Bignoli, and T.~Petersen.
\newblock {WWF (2022) Living Planet Report 2022 – Building a naturepositive
  society.}
\newblock \url{https://livingplanet.panda.org/en-US/}, 2022.

\bibitem[Bibby et~al.(1998)Bibby, Jones, and Marsden]{bibby1998bird}
Colin~J Bibby, Martin Jones, and Stuart Marsden.
\newblock \emph{Bird surveys}.
\newblock Expedition Advisory Centre London, 1998.

\bibitem[Corcoran et~al.(2021)Corcoran, Winsen, Sudholz, and
  Hamilton]{corcoran2021automated}
Evangeline Corcoran, Megan Winsen, Ashlee Sudholz, and Grant Hamilton.
\newblock Automated detection of wildlife using drones: Synthesis,
  opportunities and constraints.
\newblock \emph{Methods in Ecology and Evolution}, 12\penalty0 (6):\penalty0
  1103--1114, 2021.

\bibitem[Finn et~al.(2023)Finn, Grattarola, and Pincheira-Donoso]{finn2023more}
Catherine Finn, Florencia Grattarola, and Daniel Pincheira-Donoso.
\newblock {More losers than winners: investigating Anthropocene defaunation
  through the diversity of population trends}.
\newblock \emph{Biological Reviews}, 98\penalty0 (5):\penalty0 1732--1748,
  2023.

\bibitem[Francis et~al.(2020)Francis, Lyons, Kingsford, and
  Brandis]{francis2020counting}
Roxane~J Francis, Mitchell~B Lyons, Richard~T Kingsford, and Kate~J Brandis.
\newblock Counting mixed breeding aggregations of animal species using drones:
  Lessons from waterbirds on semi-automation.
\newblock \emph{Remote Sensing}, 12\penalty0 (7):\penalty0 1185, 2020.

\bibitem[{Graduation}(2023)]{elephantgiraffe}
{Graduation}.
\newblock {Animals Object Detection Dataset and Pre-Trained Model}.
\newblock universe.roboflow.com/graduation-nnzal/animals-kapzz, 2023.

\bibitem[Lenzi et~al.(2023)Lenzi, Barnas, ElSaid, Desell, Rockwell, and
  Ellis-Felege]{lenzi2023artificial}
Javier Lenzi, Andrew~F Barnas, Abdelrahman~A ElSaid, Travis Desell, Robert~F
  Rockwell, and Susan~N Ellis-Felege.
\newblock Artificial intelligence for automated detection of large mammals
  creates path to upscale drone surveys.
\newblock \emph{Scientific Reports}, 13\penalty0 (1):\penalty0 947, 2023.

\bibitem[{Mostafa}(2024)]{elephant}
{Mostafa}.
\newblock {06-09-16 Object Detection Dataset and Pre-Trained Model}.
\newblock {universe.roboflow.com/mostafa-xt502/06-09-16}, 2024.

\bibitem[Norouzzadeh et~al.(2018)Norouzzadeh, Nguyen, Kosmala, Swanson, Palmer,
  Packer, and Clune]{norouzzadeh2018automatically}
Mohammad~Sadegh Norouzzadeh, Anh Nguyen, Margaret Kosmala, Alexandra Swanson,
  Meredith~S Palmer, Craig Packer, and Jeff Clune.
\newblock {Automatically identifying, counting, and describing wild animals in
  camera-trap images with deep learning}.
\newblock \emph{Proceedings of the National Academy of Sciences}, 115\penalty0
  (25):\penalty0 E5716--E5725, 2018.

\bibitem[{NPS}(2023)]{approximatecounts}
{NPS}.
\newblock {Bird Surveys}.
\newblock https://www.nps.gov/maca/learn/bird-surveys.htm, 2023.

\bibitem[Pollitt et~al.(2003)Pollitt, Hall, Holloway, Hearn, Marshall,
  Musgrove, Robinson, and Cranswick]{pollitt2003wetland}
MS~Pollitt, Colette Hall, SJ~Holloway, RD~Hearn, PE~Marshall, AJ~Musgrove,
  JA~Robinson, and PA~Cranswick.
\newblock The wetland bird survey 2000--01: wildfowl and wader counts.
\newblock \emph{British Trust for Ornithology, The Wildfowl and Wetlands Trust,
  Royal Society for the Protection of Birds and Joint Nature Conservation
  Committee}, 2003.

\bibitem[Redmon et~al.(2016)Redmon, Divvala, Girshick, and
  Farhadi]{redmon2016you}
Joseph Redmon, Santosh Divvala, Ross Girshick, and Ali Farhadi.
\newblock {You only look once: Unified, real-time object detection}.
\newblock In \emph{Proceedings of the IEEE conference on computer vision and
  pattern recognition}, pages 779--788, 2016.

\bibitem[Sahin and Ozer(2021)]{sahin2021yolodrone}
Oyku Sahin and Sedat Ozer.
\newblock Yolodrone: Improved yolo architecture for object detection in drone
  images.
\newblock In \emph{2021 44th International Conference on Telecommunications and
  Signal Processing (TSP)}, pages 361--365. IEEE, 2021.

\bibitem[Schad and Fischer(2023)]{schad2023opportunities}
Lukas Schad and Julia Fischer.
\newblock Opportunities and risks in the use of drones for studying animal
  behaviour.
\newblock \emph{Methods in Ecology and Evolution}, 14\penalty0 (8):\penalty0
  1864--1872, 2023.

\bibitem[Simonyan and Zisserman(2014)]{simonyan2014very}
Karen Simonyan and Andrew Zisserman.
\newblock {Very deep convolutional networks for large-scale image recognition}.
\newblock \emph{arXiv preprint arXiv:1409.1556}, 2014.

\bibitem[{Ultralytics}(2023)]{yolov8}
{Ultralytics}.
\newblock {YOLOv8}.
\newblock https://docs.ultralytics.com/, 2023.

\bibitem[Youngflesh et~al.(2023)Youngflesh, Montgomery, Saracco, Miller,
  Guralnick, Hurlbert, Siegel, LaFrance, and
  Tingley]{youngflesh2023demographic}
Casey Youngflesh, Graham~A Montgomery, James~F Saracco, David~AW Miller,
  Robert~P Guralnick, Allen~H Hurlbert, Rodney~B Siegel, Raphael LaFrance, and
  Morgan~W Tingley.
\newblock Demographic consequences of phenological asynchrony for north
  american songbirds.
\newblock \emph{Proceedings of the National Academy of Sciences}, 120\penalty0
  (28):\penalty0 e2221961120, 2023.

\bibitem[Zhang(2023)]{zhang2023drone}
Zhengxin Zhang.
\newblock {Drone-YOLO: An efficient neural network method for target detection
  in drone images}.
\newblock \emph{Drones}, 7\penalty0 (8):\penalty0 526, 2023.

\end{thebibliography}

\end{document}